\documentclass{article}

    \usepackage[preprint]{neurips_2024}

\usepackage[utf8]{inputenc} 
\usepackage[T1]{fontenc}    
\usepackage{hyperref}       
\usepackage{url}            
\usepackage{booktabs}       
\usepackage{amsfonts}       
\usepackage{nicefrac}       
\usepackage{microtype}      
\usepackage[dvipsnames]{xcolor}

\usepackage{ulem} 
\usepackage{graphicx}
\usepackage{subfig}
\usepackage{wrapfig}

\usepackage{tcolorbox}
\usepackage{tabularray}
\usepackage{amsmath}

\usepackage{multirow}
\usepackage{colortbl}
\usepackage{hhline}
\usepackage{vcell}

\title{On the Brittle Foundations of ReAct Prompting for Agentic Large Language Models}

\author{%
  Mudit Verma* \\
  \And
  Siddhant Bhambri* \\ 
  \And 
  Subbarao Kambhampati \\ 
  \And 
  \vspace{-0.75cm}\\
  School of Computing and AI\\ Arizona State University \\ 
  \texttt{\{muditverma, sbhambr1, rao\}@asu.edu}
}

\begin{document}

\maketitle

\begin{abstract}


The reasoning abilities of Large Language Models (LLMs) remain a topic of debate. Some methods such as ReAct-based prompting, have gained popularity for claiming to enhance sequential decision-making abilities of agentic LLMs. However, it is unclear what is the source of improvement in LLM reasoning with ReAct based prompting. In this paper we examine these claims of ReAct based prompting in improving agentic LLMs for sequential decision-making. By introducing systematic variations to the input prompt we perform a sensitivity analysis along the claims of ReAct and find that the performance is minimally influenced by the "interleaving reasoning trace with action execution" or the content of the generated reasoning traces in ReAct, contrary to original claims and common usage. Instead, the performance of LLMs is driven by the similarity between input example tasks and queries, implicitly forcing the prompt designer to provide instance-specific examples which significantly increases the cognitive burden on the human. Our investigation shows that the perceived reasoning abilities of LLMs stem from the exemplar-query similarity and approximate retrieval rather than any inherent reasoning abilities.
\end{abstract}

\section{Introduction}

Large Language Models (LLMs) have seen rapid advancements specifically in Natural Language Processing and Understanding (NLP \& NLU). LLMs have unparalleled capabilities in text generation, summarization, translation, question answering to name a few. \citep{bubeck2023sparks}. Motivated by these capabilities of LLMs, there has also been a rush to look for other emergent abilities--especially for reasoning and planning. 
A popular way of enhancing LLM performance on reasoning/planning tasks has been in-context prompting or prompt-engineering \citep{sahoo2024systematic} to include instructions \citep{giray2023prompt}, syntax structure \citep{marvin2023prompt}, criticism and plan guidance with verification \citep{kambhampati2024llms} etc. Among these approaches, ReAct \citep{yao2022react} stands out which claims to improve LLM planning abilities through the use of reasoning traces interleaved with action execution given as plan guidance. While their original study includes multiple domains tasks such as question answering on HotPotQA and FEVER datasets \citep{yang2018hotpotqa, thorne2018fever}, 
in this paper we are particularly interested in 
planning tasks
where ReAct claims superior performance. 

In our initial experiments with ReAct for planning, we found that the system is overly dependent on the syntactic similarity of the example prompt and the query and is extremely brittle to minor perturbations to the input prompt. For example, when provided with an explicit set of examples of \textit{pick-and-place-object} task and asked to plan for a \textit{pick-and-place-two-objects} task, it should be trivial to generalize the solution of the examples to the queried task. Unfortunately, even such a minor variation to the original ReAct agent setup disrupts its performance.

\begin{figure}[h]
    \centering
    \includegraphics[width=\textwidth]{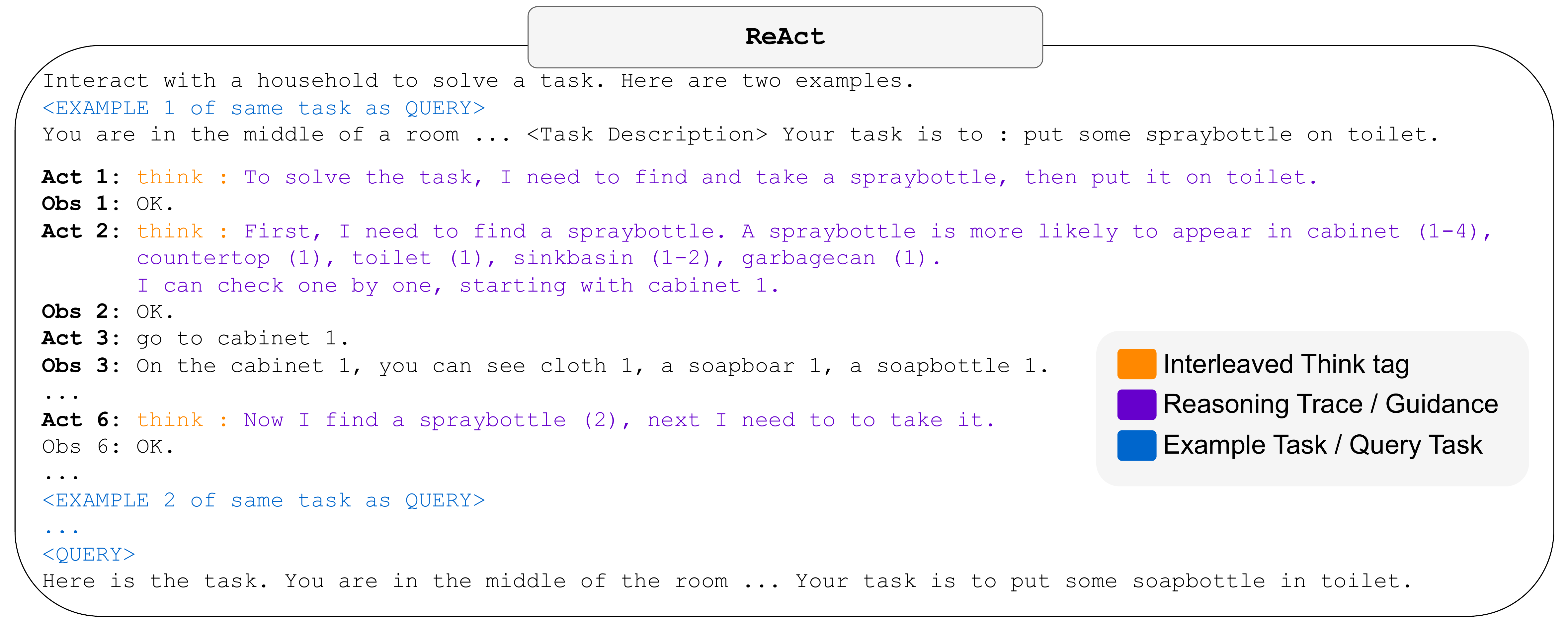}
    \caption{An example of ReAct in AlfWorld. We highlight the main components of ReAct, i.e., Interleaved reasoning and acting, the reasoning trace / plan guidance and the example and query task.}
    \label{fig:example-react}
    \vspace{-10pt}
\end{figure}

Given the seemingly wide spread adoption of ReAct methodology, the brittleness we witnessed calls for a 
systematic study of the factors 
contributing to the performance of  ReAct-based LLM Agents. Based on the claims of \citep{yao2022react}, we isolate three possible reasons for the claimed performance of ReAct framework: 1) the utility of interleaving reasoning trace during action execution, 2) the utility of providing plan guidance, and, 3) the significance of example prompt provided to the the LLM. However, the brittleness of ReAct becomes apparent when considering the variability in prompt designs. Depending on the domain and task, prompt designers may provide abstract guidance, task-specific instructions within the same domain, analogical examples, or global reasoning traces. ReAct's inability to robustly adapt to such variations underscores its limitations in handling diverse input prompts, thereby necessitating a closer examination of its design and implementation. In this work, we systematically evaluate the brittleness of ReAct by studying which potential factors contribute to its performance. This analysis is conducted through variations in input prompts to understand how a ReAct LLM Agent responds to (1) where the guidance is provided, (2) the different types and structure of this guidance, and finally, (3) on varying the resemblance of example prompt to the queried problem. We investigate the research questions : 

\textbf{RQ1:} Does the agent performance depend on interleaving reasoning trace with action execution? \\ 
\textbf{RQ2:} How does the nature of the reasoning trace or guidance information affect the performance of LLM Agents? \\ 
\textbf{RQ3:} How does the similarity between the example \textlangle problem, solution\textrangle and the query \textlangle problem, ? \textrangle, which are present in the prompt, affect LLM Agent performance? 


We conduct extensive experiments on the AlfWorld domain using various LLM Models, including GPT-3.5-turbo, GPT-3.5-instruct, GPT-4 and Claude-Opus. Through our comprehensive empirical study, answer each of the research questions above. (RQ1) We find that LLM performance in-fact improves when the reasoning trace is \textbf{not} interleaved with action execution. (RQ2) Moreover, providing weaker guidance or placebo-guidance (where the text provides no information about the task) has comparable performance to strong reasoning trace based guidance. Answers to RQ1 and RQ2 highlight that the source of ReAct LLM agent performance is \textbf{not} the interleaving aspect or the content of the reasoning trace. Finally, (RQ3) we see that variations to the example prompt such that it belongs to different task within the same domain, or has a different goal or plan attributes than the queried problem; causes the performance of ReAct-LLM Agent to plummet. 

Our findings highlight that the benefits of ReAct-based LLM Agents are present when prompt engineers can curate instance-specific examples. This approach may not scale for domains with a large number of problem instance classes, and it places an undue burden on prompt engineers to provide instance specific examples. 
Finally, our experiments call into question claims of enhanced ``emergent reasoning" of LLMs with prompt engineering efforts such as ReAct; corroborating contemporary research \citep{verma2024theory, valmeekam2024planning, stechly2024self, ullman2023large, schaeffer2023invalid, mccoy2023embers} questioning reasoning abilities of LLMs.

\section{Preliminaries}
\vspace{-0.25cm}
\label{sec:preliminaries}

\subsection{AlfWorld}
\label{subsec:prelim_alfworld}
\vspace{-0.25cm}
AlfWorld \cite{shridhar2020alfworld} is a synthetic text-based game built on top of a PDDL domain description. ReAct \cite{yao2022react} defines six tasks (or problem classes) within this domain namely - Put, Clean, Heat, Cool, Examine, and PutTwo. Each problem class consists of several problem instances, such as \textit{put a spraybottle on toilet} (see Fig. \ref{fig:example-react} is an example instance of Put class. Since AlfWorld is a partially observable environment, each of these problem instances can be solved by navigating and interacting with the environment simulator via text actions. For example, this task can be solved by the following actions- \texttt{go to cabinet 2, take spraybottle 2 from cabinet 2, go to toilet 1, put spraybottle 2 in/on toilet 1}.

\subsection{ReAct}
\vspace{-0.25cm}
\label{subsec:prelim_react}

The \textit{think} action tag provided by ReAct is claimed to comprise of \textbf{Re}asoning + \textbf{Act}ion trace that is provided in the solution for the example problems as part of the prompt. During execution, when the LLM agent is queried, the expectation is that it can generate a \textit{think} action tag for the queried problem instance that is semantically similar to the one provided for the examples in the prompt.

\textbf{Location of \textit{THINK} tag}
In ReAct, the integration of the \textit{think} tag within actions serves to expand the action space. This allows the language model (LLM) agent to execute a \textit{think} action, prompting an \textit{`OK'} response. Through analysis of example prompts in ReAct experiments, we identify various instances of the \texttt{think} action. Typically, it appears after stating the problem instance, reiterating the task, and providing problem-specific guidance. However, the authors offer no structured guidelines for its implementation, placement, or guidance. This observation aligns with feedback from reviewers \citep{openreviewReactComments} citing inconsistencies in the prompting format.

\textbf{Content of \textit{THINK} tag}
In ReAct, the \textit{think} action consistently provides the decision-making agent with success-oriented guidance for task completion. For instance, upon encountering a \texttt{spraybottle}, the prompt might include: \texttt{think: Now I find a spraybottle (2). Next, I need to take it.} This guidance exposes forthcoming actions and sub-tasks for the agent.

\textbf{Few shot \textit{EXAMPLE}s}
In the AlfWorld domain (wihch is a PDDL domain), ReAct authors \citep{yao2022react} classify six problem classes or tasks: \texttt{Put, Clean, Heat, Cool, Examine, PutTwo}. Despite representing different tasks, they share the same environment dynamics and action space, allowing for very similar execution trace. For instance, a \texttt{Heat} task might involve \texttt{Put}ting an item into a microwave. In ReAct experiments, authors provide two example problem-solution pairs (referred to as exemplars in our work) before querying the LLM agent with a problem instance. Authors force ReAct agent to use examples and queries belonging to the same problem class without motivating this design decision. However, the queried problem may differ in objects or locations from the exemplars.

\label{sec:methodology}

\section{Related Work}
\label{sec:related_work}
\vspace{-0.25cm}

Large Language Models have been shown to be successful in a plethora of natural language tasks \citep{kocon2023chatgpt,gilardi2023chatgpt,zhu2023can,bubeck2023sparks,bhattacharjee2024towards}. However, there are two schools of thought when it comes to utilizing off-the-shelf LLMs for planning and reasoning tasks. Works such as Chain of Thought, ReAct, and others that followed \citep{wei2022chain,yao2023beyond,long2023large,yao2024tree,besta2024graph,fu2024preact,aksitov2023rest}, have argued about the reasoning abilities of LLMs by proposing prompting methods. On the other hand, \citep{valmeekam2024planning,stechly2024self} have refuted these claims by showing the inability of LLMs to solve deterministic planning and classical reasoning problems.

In particular, for investigating the use of LLMs in solving decision making problems, AlfWorld \citep{shridhar2020alfworld} is a popular domain that was originally proposed for training text-based Reinforcement Learning agents. Lately, works such as ReAct, Reflexion, and their other variants \citep{yao2022react,shinn2023reflexion} have argued on the prowess of LLMs' reasoning abilities on AlfWorld. Furthermore, there have been several extensions to ReAct that boost their generalization abilities across more domains including multi-modal domains \citep{yang2023mm,castrejon2024hammr}, autonomous vehicles \citep{cui2024receive}, table question answering \citep{zhang2023reactable}, etc. While the effectiveness of ReAct is celebrated across different areas, these works only depend on anthropomorphization of LLMs for using ReAct based prompting with no justification on the source of improvement in performance. This motivates our work in investigating the components of ReAct with respect to sequential decision-making problems and analyzing the role each component plays.

\vspace{-0.25cm}
\section{Brittleness of ReAct based Agentic LLMs}
\vspace{-0.25cm}

We examine the claims of ReAct to understand the performance of ReAct-based LLM agents. It is crucial to assess whether ReAct's fundamental assertions hold, particularly in sequential decision making. As outlined in Section \ref{sec:preliminaries}, ReAct comprises three main components: interleaving the \texttt{think} tag with actions, plan guidance after the \texttt{think} tag, and the selection of exemplar problems for LLM prompts. We perform a sensitvity analysis by proposing alternatives along these three dimensions. The subsequent sections explore the design of exemplar prompt variations to investigate our research questions concerning the claims of ReAct. Each variation modifies the base ReAct exemplar prompts.

\begin{figure}[t]
    \centering
    \includegraphics[width=\textwidth]{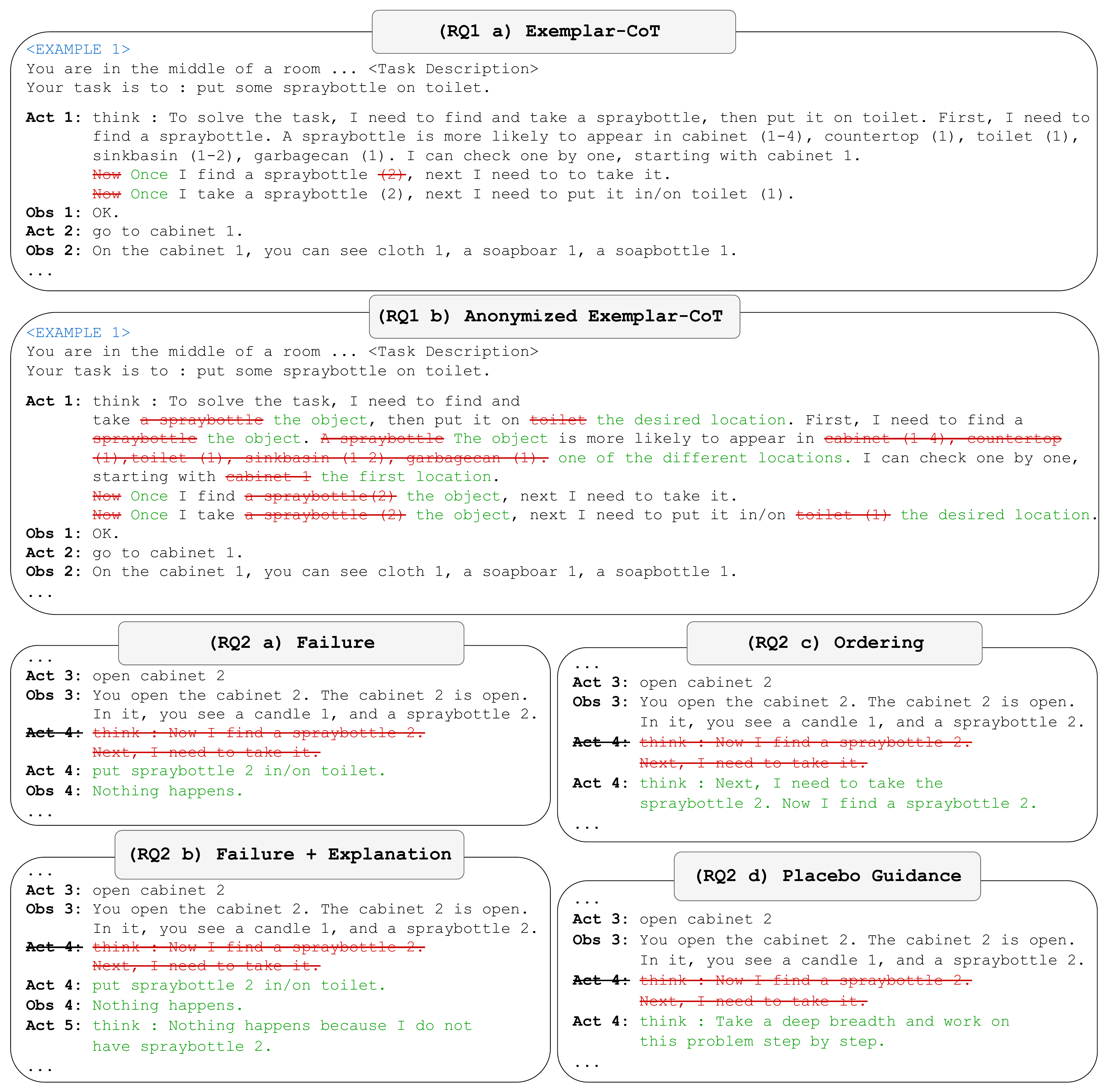}
    \caption{Example of prompt variations considered for RQ1 and RQ2.}
    \label{fig:examples-h1-h2}
\end{figure}

\subsection{RQ1 : Interleaving \textit{think}ing with acting}
\label{subsec:method_think_location}
\vspace{-0.25cm}

\textit{Does the agent performance depend on interleaving reasoning trace with action execution?}

To answer this research question, we propose collating the guidance information contained within the multiple \textit{think} tags present in the examples of the input prompt into a single \textit{think} tag appended after the example problem is specified. This approach can be interpreted as Chain-of-Thought \cite{kojima2022large, wei2022chain}, where guidance information is generated before action execution.

\textbf{Variation 1: Exemplar-based CoT}
AlfWorld is a partially observable environment where an agent can only observe objects after reaching that location. Hence, we remove specific location and object identifiers to modify the \textit{think} actions that are originally interleaved with other actions in the environment. For example, we rewrite it as: \texttt{think: {\color{Mahogany}\sout{Now}} \textcolor{PineGreen}{Once} I find a spraybottle {\color{Mahogany}\sout{(2)}}, next I need to take it.}. Finally, we append all the \textit{think} actions together at the beginning of the example problem.
\textbf{Intuition:} Problem-specific guidance for a sequential decision-making agent can be given step-by-step (as in ReAct) or all at once.

\textbf{Variation 2: Anonymized Exemplar-CoT }
We take one step further and modify the \textit{think} tag to remove references to specific locations and objects, making it more general. For example, \texttt{think: To solve the task, I need to find and take {\color{Mahogany}\sout{a spraybottle}} \textcolor{PineGreen}{the object}, then put it on {\color{Mahogany}\sout{toilet}} \textcolor{PineGreen}{the desired location}}.
\textbf{Intuition:} Exemplars can be made more general by providing abstract guidance and exploting LLMs ability to identify necessary semantic entity relations.

\vspace{-0.25cm}
\subsection{RQ2 : Plan Guidance following \textit{think} tag}\
\vspace{-0.2cm}
\label{subsec:method_think_content}
\textit{How does the nature of the reasoning trace or guidance information affect the performance of LLM?}

ReAct claims to use reasoning trace as the guidance information following the \textit{think} tag. For instance, in ReAct \cite{} thoughts are to (1) decompose the goal (2) track subgoal completion (3) determine the next subgoal and (4) reason via common-sense where to find and object and what to do with it. It is, however, unclear what is the motivation to use these as the reasoning trace. The potential anthropomorphization of large language models (LLMs) may suggest that their thought processes are similar to the abstract plans humans make, and that they must be prompted in the same manner. However, it is unclear why this assumption should hold true. Alternatives can be, we can prompt the LLM to reflect on past failures and provide possible explanations (hindsight-guidance) or We can substitute task-relevant guidance with placebo-guidance by using "magic incantations". 


\textbf{Variation 1: Failure}
From the example prompts used in ReAct, we note that none of the examples for any task consist of invalid actions. We inject two invalid actions in the execution trace : the first that attempts to execute the action pertinent to the task (such as \texttt{put spraybottle 2 in/on toilet}) when not possible and, second, executes some other invalid action. We include the expected simulator response, \texttt{Nothing happens.}, when invalid actions are taken. \textbf{Intuition:} Reasoning trace can be about \textit{what to do} such as subgoals of the future, or \textit{what not to do} such as mistakes in hindsight. This should be weaker guidance than in base ReAct as the exemplars do not point out what to do next.

\textbf{Variation 2: Failure + Explanation}
We place \textit{think} actions after invalid actions injected in Failure Variation which consist of explanations for the failure such as \textcolor{PineGreen}{\texttt{think: Nothing happens because I do not have a spraybottle 2}}. \textbf{Intuition:} We can augment pointing out mistakes in hindsight with explanations to avoid similar failures. This is stronger guidance signal than Failure, however, the exemplars still not provide information on what to do next.

\textbf{Variation 3: Guidance Ordering}
LLMs are known to be susceptible to minor syntactic perturbations to inputs. We test whether it is true for guidance information given as prompt as well. We identify chain of subtasks in a reasoning trace $S_1 \rightarrow S_2 \cdots S_n$ and reverse it to be $S_n \rightarrow S_{n-1} \cdots S_1$. For instance, \textcolor{Mahogany}{\texttt{think: Now I find a spraybottle (2). Next, I need to take it.}} becomes \textcolor{PineGreen}{\texttt{think: Next, I need to take the spraybottle (2). Now I find a spraybottle (2)}}. \textbf{Intuition:} LLM agent should be invariant to the syntax of reasoning trace if the semantic information is preserved. This does not change the reasoning trace from the perspective of information content.

\textbf{Variation 4: Placebo Guidance}
It is unclear to what extent LLM agent uses the supposed helpful thoughts for the decision making task. In this variation we replace \textit{think} tag guidance with a placebo thought that does not contain any task relevant information, but has been widely used as prompt engineering trick \citep{kojima2022large}. \textbf{Intuition:} According to claims of ReAct, we expect the performance to get worse when the guidance does not have any information useful for task success.

\vspace{-0.25cm}
\subsection{RQ3 : Similarity between \textit{EXAMPLE}s and \textit{QUERY}}
\vspace{-0.25cm}
\label{subsec:method_example_similarity}

\textit{How does the similarity between the example \textlangle problem, solution\textrangle and the query \textlangle problem, ? \textrangle, which are present in the prompt, affect LLM Agent performance?}

RQ3 investigates the role of example similarity to the query in LLM agent's performance. Establishing problem similarity can be challenging, especially where minor variations to the problem can have varied interpretations (such as an analogy to a different task altogether). Our work explores this challenge in a systematic way. During example prompt construction,  prompt designers may use synonyms to refer to objects (\texttt{Domain}), come up with examples where the agent task is the same as query but the goals are different (\texttt{Instance}), or provide optimal solutions as the examples (\texttt{Optimal}) preventing LLM to obtain information regarding exploration strategy. Furthermore, given that the domain has the same underlying action dynamics and that the tasks reuse several actions, prompt designers may choose to provide query specific example prompts (as in base \texttt{ReAct}), provide one of a different task and one of the same task (\texttt{One}), provide both examples to be of a different task (\texttt{Both}), or take an exhaustive approach and provide one example of all tasks (\texttt{All}).

\textbf{Variation 1: Synonyms - (\texttt{Domain})}
For this variation, we replace the object and location names in the example prompts with their synonyms. For example, \texttt{{\color{Mahogany}\sout{spraybottle}} $\rightarrow$ \textcolor{PineGreen}{aerosolbottle}}, \texttt{{\color{Mahogany}\sout{cabinet}} $\rightarrow$ \textcolor{PineGreen}{cupboard}}, and, \texttt{{\color{Mahogany}\sout{microwave}} $\rightarrow$ \textcolor{PineGreen}{oven}}. We make 36 such changes to object and location names across all the examples. Note that the object names / location are unchanged for the problem query and subsequent interaction with the simulator.
\textbf{Intuition:} Exemplar guidance maybe specified with alternate synonymous object and location names. Reasoning agents should be invariant to variable name substitution for closed world dynamics such as PDDL based AlfWorld.


\textbf{Variation 2: Problem Instance-level - \texttt{Instance}}
We inject instance-level changes to the examples provided in the prompts. Recall that we are updating the base ReAct's prompts, where the exemplar tasks are same as the query. We change the goal location in exemplar problem to ensure that it does not match with any of the goal locations in query problem. Moreover, we add repetitive yet futile actions in the exemplar execution trace which does not effect the solution. \textbf{Intuition:} Ensuring a different goal location in exemplar from the queried problem is a natural usecase. Moreover, exemplars may contain arbitrary exploration strategies such as action repetition \citep{sharma2017learning}. By ReAct's claims, LLM agent performance should not be affected.

\textbf{Variation 3: Problem Level - \texttt{Both, One, All}}
Recall that the environment dynamics for all the tasks are the same. In fact, several tasks subsume the use of our tasks such as \texttt{Heat} requires the agent to \texttt{Put} an food in the microwave. In general, all the tasks share a large portion of actions (such as exploring cabinets and locations, picking objects etc.). Motivated by how tight relationship of these tasks we come up with three variations. First, \texttt{One}, uses one exemplar of an arbitrarily picked task and the other exemplar of the same task as the query. Second, \texttt{Both}, uses both exemplars from an arbitrarily picked task. Finally, \texttt{All}, uses a total of six exemplars (this is the only variation where we provide more than the standard two examples as in ReAct) corresponding to each task under consideration. Remember, this includes the query task which is always present at the end in the input prompt. \textbf{Intuition:} With a very similar action execution trace (such as exploration, picking and placing objects) across tasks, and shared dynamics, LLM agent should be minimally affected by the use of exemplars of a different task.

\textbf{Variation 4: Exploration Strategy - \texttt{Optimal}}
As noted before, ReAct does not explain the choice of exemplars used. An important ingredient to the exemplars is the exploration strategy used. In this variation we provide exemplars which serendipitously take the optimal actions (as if the environment were fully observable) and therefore the example plan is the shortest possible. \textbf{Intuition:} Exploration strategy exposed in exemplars (that too for the same problem task) should not impact ReAct's performance if the LLM agent is reasoning instead of retrieval (or pattern matching).

\vspace{-0.25cm}
\section{Results}
\vspace{-0.25cm}

\begin{table}
\centering
\caption{Average Success \% of LLM on Base ReAct and prompt variations for RQ1 and RQ2 on six AlfWorld tasks.}
\begin{tabular}{c|c|cc|cccc} 
\hline
\label{table:h1-h2}
\multirow{2}{*}{\textbf{Model / Prompt}} & \multirow{2}{*}{\textbf{Base}} & \multicolumn{2}{c|}{\textbf{H1}} & \multicolumn{4}{c}{\textbf{H2}} \\ 
\cline{3-8}
 &  & \textbf{CoT} & \textbf{Anon. CoT} & \textbf{Magic} & \textbf{Order} & \textbf{Failure} & \textbf{Explanation} \\ 
\hline
\vcell{GPT-3.5-Turbo} & \vcell{27.6} & {\cellcolor[rgb]{0.29,1,0.278}}\vcell{46.6} & {\cellcolor[rgb]{0.459,1,0.522}}\vcell{41} & {\cellcolor[rgb]{0.659,1,0.718}}\vcell{30} & {\cellcolor[rgb]{0.8,1,0.851}}\vcell{28.3} & {\cellcolor[rgb]{0.345,0.996,0.353}}\vcell{43.3} & {\cellcolor[rgb]{0.278,1,0.365}}\vcell{41.6} \\[-\rowheight]
\printcellbottom & \printcellmiddle & {\cellcolor[rgb]{0.29,1,0.278}}\printcellmiddle & {\cellcolor[rgb]{0.459,1,0.522}}\printcellmiddle & {\cellcolor[rgb]{0.659,1,0.718}}\printcellmiddle & {\cellcolor[rgb]{0.8,1,0.851}}\printcellmiddle & {\cellcolor[rgb]{0.345,0.996,0.353}}\printcellmiddle & {\cellcolor[rgb]{0.278,1,0.365}}\printcellmiddle \\
GPT-3.5-Instruct & 44.7 & {\cellcolor[rgb]{0.161,1,0.298}}61.9 & {\cellcolor[rgb]{0.361,1,0.373}}50.7 & {\cellcolor[rgb]{1,0.898,0.898}}41 & {\cellcolor[rgb]{1,0.941,0.941}}42.5 & {\cellcolor[rgb]{0.745,0.996,0.796}}47 & 44.7 \\
GPT-4 & 23.3 & {\cellcolor[rgb]{0.141,1,0.282}}43.3 & {\cellcolor[rgb]{0.486,0.996,0.569}}33.3 & {\cellcolor[rgb]{0.322,1,0.431}}36.6 & {\cellcolor[rgb]{0.58,1,0.686}}30 & {\cellcolor[rgb]{0,1,0.165}}50 & {\cellcolor[rgb]{0.322,1,0.431}}36.6 \\
Claude-Opus & 56.6 & {\cellcolor[rgb]{1,0.898,0.898}}50 & {\cellcolor[rgb]{0.996,0.765,0.765}}46.6 & {\cellcolor[rgb]{0.996,0.624,0.624}}30 & {\cellcolor[rgb]{1,0.898,0.898}}50 & {\cellcolor[rgb]{1,0.961,0.961}}53.3 & {\cellcolor[rgb]{0.996,0.545,0.545}}30 \\
\hline
\end{tabular}
\end{table}

In the following sub-sections, we will answer our RQs through sensitivity analysis using the proposed prompt variations along three dimensions, the location of the \textit{think} tag, the content of the \textit{think} tag, and the similarity between exemplars and queried problems. All the variations modify the base ReAct prompts and we do not present a cross between variations unless otherwise noted. We use GPT-3.5-Turbo, GPT-3.5-Instruct, GPT-4, and Claude-Opus which are all newer models than those benchmarked in ReAct \citep{yao2022react}. As noted, we use AlfWorld domain consistent with the setup in \citep{yao2022react}. GPT3.5(Turbo, Instruct) results are on 134 instances across six tasks, GPT-4/Claude-Opus on 60 instances (10 for each task) due to cost considerations.

\subsection{Utility of interleaving reasoning trace with action execution}
\label{subsec:results_location}

\begin{figure}[t]
    \centering
    \includegraphics[width=\textwidth]{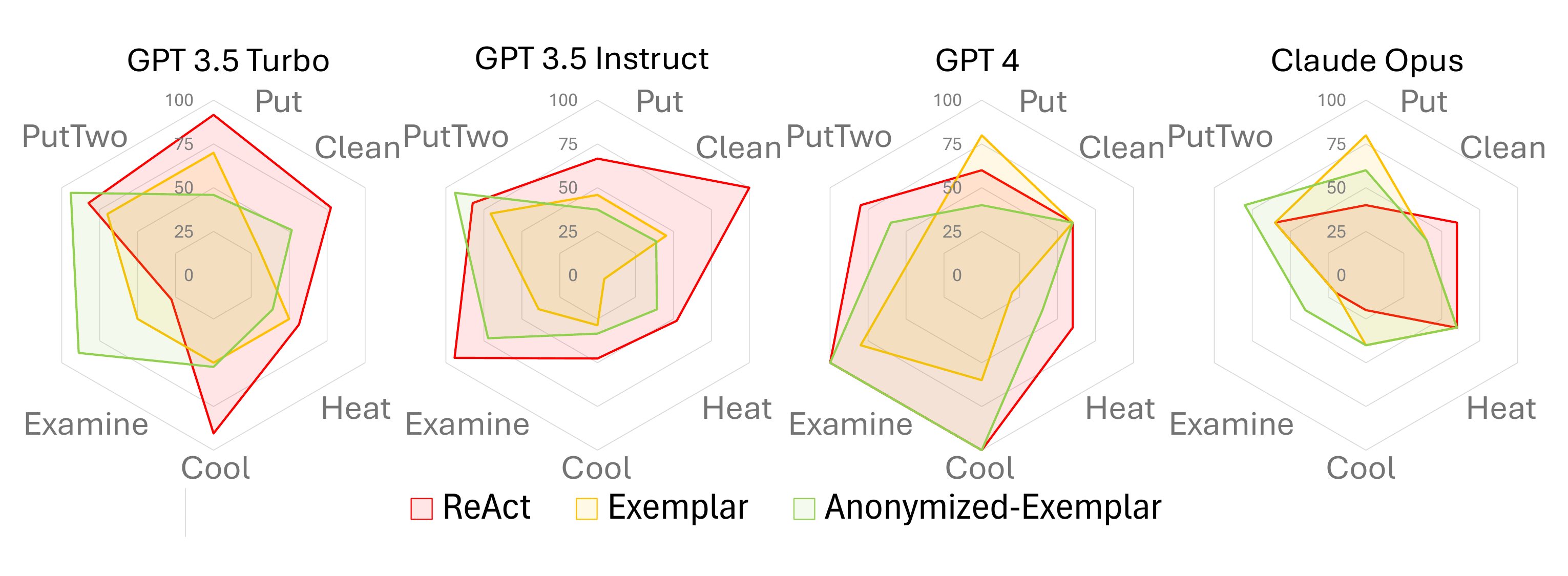}
    \caption{The radar chart shows the \textbf{failure rates} of various LLMs with different ReAct-based prompt settings for RQ1 (Base React, Global, Anonymized) across six Alfworld tasks (hexagon vertices). Higher values / Larger shaded region indicate worse performance.}
    \label{fig:results_h1}
\end{figure}

From Table \ref{table:h1-h2}(RQ1) note that the exemplar CoT and the anonymized exemplar CoT performs significantly better than base ReAct for all GPT-X family of models. Moreover, the performance dips slightly for Claude-Opus along these variations. From Fig. \ref{fig:results_h1} (larger area represents worse performance), we observe that base ReAct consistently performs worse in most of the tasks. This refutes ReAct's first claim on the importance of interleaving reasoning trace generation with action execution. Even in the case of the Claude where there is a slight dip in performance, the models seems to be performing at reasonably high success rate which questions the importance of interleaved reasoning and action execution.

\vspace{-0.25cm}
\subsection{Utility of Guidance Information following \textit{think} tag}
\label{subsec:results_content}

\begin{figure}
    \centering
    \includegraphics[width=\textwidth]{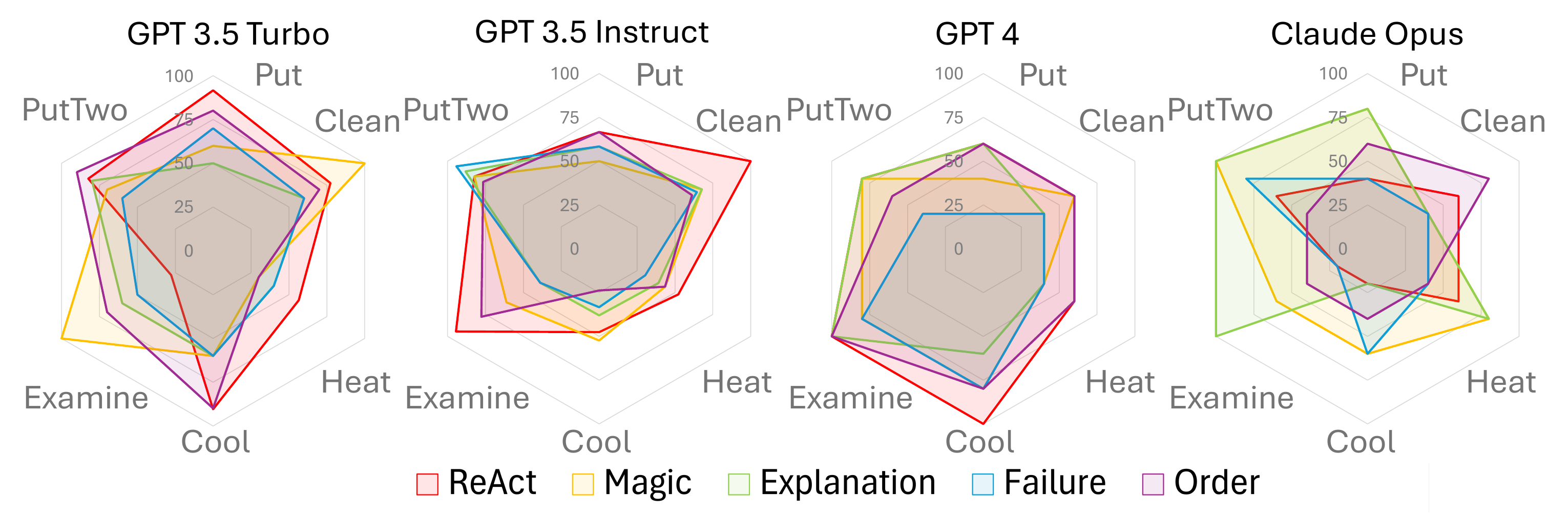}
    \caption{The radar chart shows the \textbf{failure rates} of various LLMs with different ReAct-based prompt settings for RQ2 (Base React, Magic, Failure, Failure+Explanation, Ordering) across six Alfworld tasks (hexagon vertices). Higher values / Larger shaded region indicate worse performance.}
    \label{fig:results_h2}
\end{figure}

Recall that reasoning trace guidance pertains to the prospective actions or behaviors an agent should execute (foresight guidance). This type of guidance is more informative compared to other variations, such as hindsight guidance, which focuses on past errors without providing future solution steps, and placebo guidance, which is entirely unrelated to the task. ReAct claims that reasoning trace is crucial for LLM agent performance, which would predict a decline in performance with hindsight guidance and a collapse with placebo guidance. In contrast, our findings in Table \ref{table:h1-h2} indicate that hindsight guidance (\texttt{Failure, Explanation}) actually improve the performance of the GPT family of models. The Claude-Opus model's performance remains stable with hindsight (\texttt{Failure}) guidance and declines with placebo guidance. Figure \ref{fig:results_h2} illustrates these models' performance across six AlfWorld tasks and variations, showing that the performance of LLMs either improved or remained consistent when provided with weaker or irrelevant guidance information. This refutes ReAct's claim that task-specific reasoning trace is the source of LLM agent performance. Our argument that LLM agent's performance is only slightly affected by the reasoning trace explains the indifference to ordering perturbation as well. If the LLM is not utilizing the reasoning trace for decision making, change in ordering would not affect the agent's performance. Finally, contrary to the general perception that better GPT models would improve over reasoning, we find that GPT-4's performance is the worst among GPT-X family further highlighting the brittleness of claims of ReAct. 


\vspace{-0.25cm}
\subsection{Utility of Exemplar similarity to Query task}
\vspace{-0.25cm}
\label{subsec:results_example}

\begin{table}
\centering
\caption{Average Success \% of LLM on Base ReAct and prompt variations for RQ3 on six AlfWorld tasks. OC: Out of context limit}
\begin{tabular}{c|c|cccccc} 
\hline
\label{table:h3}
\multirow{2}{*}{\textbf{Model / Prompt}} & \multirow{2}{*}{\textbf{Base}} & \multicolumn{6}{c}{\textbf{RQ3}} \\ 
\cline{3-8}
 &  & \textbf{Domain} & \textbf{Instance} & \textbf{Optimal} & \textbf{All} & \textbf{One} & \textbf{Both} \\ 
\hline
\vcell{GPT-3.5-Turbo} & \vcell{27.6} & {\cellcolor[rgb]{0.996,0.165,0.165}}\vcell{\textcolor{white}{1.6}} & {\cellcolor[rgb]{0.847,1,0.78}}\vcell{30} & {\cellcolor[rgb]{0.961,0.659,0.659}}\vcell{20.1} & {\cellcolor[rgb]{0.71,0.984,0.675}}\vcell{32} & {\cellcolor[rgb]{0.941,1,0.945}}\vcell{28.3} & {\cellcolor[rgb]{0.996,0.165,0.165}}\vcell{\textcolor{white}{1.6}} \\[-\rowheight]
\printcellbottom & \printcellmiddle & {\cellcolor[rgb]{0.996,0.165,0.165}}\printcellmiddle & {\cellcolor[rgb]{0.847,1,0.78}}\printcellmiddle & {\cellcolor[rgb]{0.961,0.659,0.659}}\printcellmiddle & {\cellcolor[rgb]{0.71,0.984,0.675}}\printcellmiddle & {\cellcolor[rgb]{0.941,1,0.945}}\printcellmiddle & {\cellcolor[rgb]{0.996,0.165,0.165}}\printcellmiddle \\
\vcell{GPT-3.5-Instruct} & \vcell{44.7} & {\cellcolor[rgb]{0.933,0.992,0.929}}\vcell{47.6} & {\cellcolor[rgb]{1,0.78,0.78}}\vcell{42.5} & {\cellcolor[rgb]{0.988,0.631,0.631}}\vcell{39.5} & \vcell{OC} & {\cellcolor[rgb]{0.984,0.514,0.514}}\vcell{17.9} & {\cellcolor[rgb]{0.976,0.204,0.204}}\vcell{\textcolor{white}{5.2}} \\[-\rowheight]
\printcellbottom & \printcellmiddle & {\cellcolor[rgb]{0.933,0.992,0.929}}\printcellmiddle & {\cellcolor[rgb]{1,0.78,0.78}}\printcellmiddle & {\cellcolor[rgb]{0.988,0.631,0.631}}\printcellmiddle & \printcellmiddle & {\cellcolor[rgb]{0.984,0.514,0.514}}\printcellmiddle & {\cellcolor[rgb]{0.976,0.204,0.204}}\printcellmiddle \\
\vcell{GPT-4} & \vcell{23.3} & {\cellcolor[rgb]{1,0.58,0.58}}\vcell{13.3} & \vcell{23.3} & {\cellcolor[rgb]{0.016,0.965,0.078}}\vcell{50} & \vcell{23.3} & {\cellcolor[rgb]{0.996,0.706,0.706}}\vcell{16.6} & {\cellcolor{red}}\vcell{\textcolor{white}{0}} \\[-\rowheight]
\printcellbottom & \printcellmiddle & {\cellcolor[rgb]{1,0.58,0.58}}\printcellmiddle & \printcellmiddle & {\cellcolor[rgb]{0.016,0.965,0.078}}\printcellmiddle & \printcellmiddle & {\cellcolor[rgb]{0.996,0.706,0.706}}\printcellmiddle & {\cellcolor{red}}\printcellmiddle \\
\vcell{Claude-Opus} & \vcell{56.6} & {\cellcolor[rgb]{0.957,0.722,0.722}}\vcell{50} & {\cellcolor[rgb]{1,0.678,0.678}}\vcell{46.6} & {\cellcolor[rgb]{0.992,0.525,0.525}}\vcell{43.3} & {\cellcolor[rgb]{0.957,0.722,0.722}}\vcell{50} & {\cellcolor[rgb]{0.831,0.992,0.827}}\vcell{60} & {\cellcolor{red}}\vcell{\textcolor{white}{6.6}} \\[-\rowheight]
\printcellbottom & \printcellmiddle & {\cellcolor[rgb]{0.957,0.722,0.722}}\printcellmiddle & {\cellcolor[rgb]{1,0.678,0.678}}\printcellmiddle & {\cellcolor[rgb]{0.992,0.525,0.525}}\printcellmiddle & {\cellcolor[rgb]{0.957,0.722,0.722}}\printcellmiddle & {\cellcolor[rgb]{0.831,0.992,0.827}}\printcellmiddle & {\cellcolor{red}}\printcellmiddle \\
\hline
\end{tabular}
\end{table}

\begin{figure}[t]
    \centering
    \includegraphics[width=\textwidth]{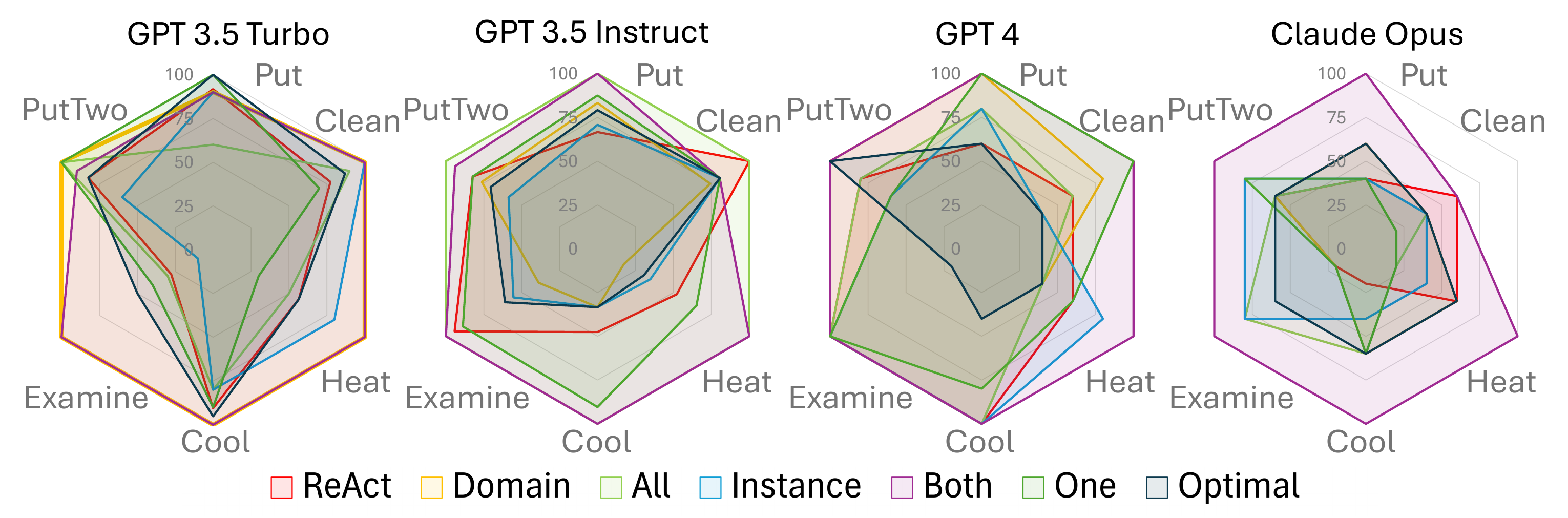}
    \caption{The radar chart shows the \textbf{failure rates} of various LLMs with different ReAct-based prompt settings for RQ3 (Base React, Domain, Instance, All, Both, One) across six Alfworld tasks (hexagon vertices). Higher values / Larger shaded region indicate worse performance.}
    \label{fig:result_h3}
\end{figure}

Intuitively, the similarity of \texttt{Domain} examples is closest with base ReAct, followed by \texttt{Instance} and \texttt{Optimal} variations. Finally, \texttt{All} contains an overload of information followed by \texttt{One} and \texttt{Both} which has the same action space but uses different tasks as exemplars. Recall that AlfWorld being a PDDL domain has a shared environment dynamics across all tasks with upto ~80\% of actions shared across execution traces. While ReAct does not investigate impact of varied exemplars, given the popular usage one expects LLMs to be robust to such changes especially in a common-sense household domain. Table \ref{table:h3} shows the severe brittleness of ReAct based LLM agent to even minor variations (such as \texttt{Domain, Instance}). Specifically, performance of GPT-3.5-Turbo and GPT-4 plumments for \texttt{Domain}. Claude-Opus which was more robust in RQ1, RQ2, is also impacted severely by \texttt{Domain, Instance} variations. Furthermore, when we do not expose the exploration strategy and only provide Optimal exemplars, the performance of LLM agents further drops (except in GPT4). 

Overloading the LLMs with more exemplars \texttt{All} does not impact its performance. We posit, this is because the query-task exemplar is still part of the large input prompt. Among the two exemplars, as provided in ReAct, when one of them is of a different task (\texttt{One}) then the performance significantly reduces for LLMs. When both of the exemplars are of a different task then the performance collapses to single digit success rates for all the models. This is a key result of this work highlighting the severe dependence of LLMs on the similarity of the exemplars to the query task. Through sensitivity analysis using our RQ3 variations we could find parts of the input (the task similarity of the exemplar with query) which is the source of ReAct performance. Essentially, the LLM is mimicking / performing approximate retrieval from the context presented to it. Moreover, our results corroborates the line of research that points out the inability of LLMs to reason or plan.

\textbf{Unrolling and Subtask Similarity} We perform additional experiments where the query task is to essentially repeat the task in the exemplar (\texttt{Unrolling}). For instance, the exemplar is \texttt{Put} and the query is \texttt{PutTwo} to put two objects at given location. In this case, the LLM has to unroll the given advice and repeat exemplar task execution to solve the query. The success rate of GPT-3.5-Instruct (the best performing GPT model in our experiments) drops down from 52\% to 9\%. Similarly, we experiment with a \texttt{Subtask Similarity} variation where the exemplar task subsumes execution of the query task. For instance, the \texttt{Heat} task requires the agent to pick and place object in the microwave (which is an instantiation of \texttt{Put} task). One would expect that \texttt{Heat} is a good exemplar for \texttt{Put}, however, the performance of GPT-3.5-instruct model goes from 18\% to 0\% in this case. These results further underscore the brittleness and the need for instance-specific exemplars in ReAct.

\textbf{Thought operationalization ability of LLMs} Given the free form nature of thought generation and arbitrary nature of thought (about subtask, common-sense next steps etc.), checking whether the generated thoughts are in-fact reasonable is a challenging problem. For completeness, we find that 40\% of the times after generation of a \textit{think} tag, subsequent environment action taken by the LLM was invalid (for GPT-3.5-instruct). It is much higher (~80\% for GPT-3.5-Turbo, ~90\% for Claude-Haiku) for weaker LLM models. This further highlights the inability of LLMs to operationalize its generated thought as also seen in \citep{roy2024flap}. From manual inspection we find that the typical thoughts would enlist all possible locations as next locations to visit for most of the tasks. As demonstrated in Section \ref{subsec:results_content}, the performance of LLMs actually decreases when provided with foresight guidance, as seen with the base ReAct model. A detailed investigation into the validity of the generated reasoning traces is beyond the scope of this work and is suggested as future research.

\section{Conclusion}
\vspace{-0.25cm}

ReAct based prompt engineering methods have been claimed to improve planning abilities of Large Language Models. 
In this study, we critically examine ReAct along three dimensions, informed by its claims and our hypotheses regarding its performance sources. Contrary to ReAct's claims, our findings reveal that its performance is \textbf{neither} due to interleaving reasoning trace and guidance information generation with action execution, \textbf{nor} due to the specific nature of the guidance information. Instead, we identify that the true source of LLM performance in sequential decision-making tasks, such as AlfWorld, is the high degree of similarity between exemplar problems (few-shot) and the query task. We also showed that ReAct is susceptible to trivial variations in exemplar prompts (such as with the use of synonyms, or \texttt{Unrolling} and \texttt{Subtask Similarity} cases). Our findings caution against an uncritical adoption of ReAct-style frameworks for their putative abilities to enhance performance in domains requiring planning.   


\section{Broader Impact \& Limitations}
\vspace{-0.25cm}
This work highlights the need for critical examination of prompt-engineering methods which claim emergent abilities of LLMs. Absence or presence of such abilities can have a large impact on several sectors such as economics, healthcare, transport to name a few. We hope highlighting the brittleness of popularly used technique such as ReAct will improve experimentation standards for the agentic LLM community. 

We deliberately confined our experiments to be deep along a domain considered common-sense planning domain by the contemporary research community. Moreover, we restricted our discussion to sequential decision making problem of AlfWorld and admit that other prompting solutions to reasoning problems must be re-evaluated with similar scrutiny. While this can be seen as a potential limitation, our examination of ReAct can be easily translated to other domains of interest by the consumer of this research.

\newpage
\bibliographystyle{plainnat}
\bibliography{refs}

\newpage

\appendix

\section{Resources Used}

In this work we leverage OpenAI API and Claude API for prompting the Language Models. We use \texttt{gpt-4-0613} for GPT4, \texttt{gpt-3.5-turbo-0125}, \texttt{gpt-3.5-turbo-instruct}, \texttt{claude-3-opus-20240229}, \texttt{claude-3-sonnet-20240229} and \texttt{claude-3-haiku-20240307} for all our experimentation in April-May 2024. ReAct and corresponding experiments use approximately 14M input tokens (due to repeated prompting after each action execution) and 150K output tokens for 134 problem instances as used by ReAct. 

\section{Additional Considerations}

\subsection{Failure Rates}
\begin{figure}[h]
    \centering
    \includegraphics[width=0.4\textwidth]{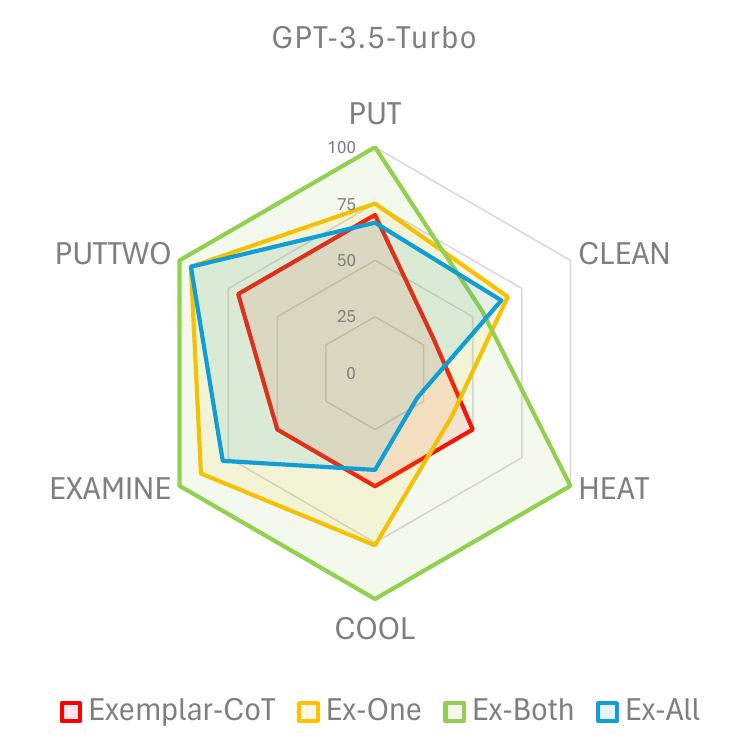}
    \caption{The radar chart shows the failure rates of GPT-3.5-Turbo with our RQ3 variants on ReAct across six Alfworld tasks. Higher values / Larger shaded region indicate worse performance.}
    \label{fig:results_global}
\end{figure}

We report failure rates in the radar chart as in Figs. \ref{fig:results_h1}, \ref{fig:results_h2}, \ref{fig:result_h3} and \ref{fig:results_global} instead of success rates. We attempted to visualize the severe brittleness given by the larger area of the shaded region. Since, for various of our RQ variations the LLMs performance was very low, we decided to report failure rate given as (100 - Success Rate \%) instead. 

\subsection{Performance of Claude-Haiku}
We skip on mentioning the performance of Claude-Haiku, since it was not able to generate syntactically correct actions for any of the instances. We found that following our instruction to generate specific actions as in the exemplar was difficult. We improved the prompt to have specific instructions for generating actions (See \ref{app:building-prompt} ) but it did not yield any improvements for Claude-Haiku. However, the instruction did help with Claude-Sonnet and Claude-Opus. We find that Claude-Sonnet follows a similar pattern as GPT-3.5-Instruct as presented in our results, and decided to focus ourselves on the strongest/largest Claude model (Claude-Opus) for our evaluation.

\subsection{Extension to other Models}
We are in the process of experimenting with GPT-4o and Google Gemini models, APIs for which were released in May 2024 which does not allow enough time for thorough and verified evaluation before the conference submission. For completeness, however, we will experiment with these APIs as they become accessible and append our results. 

\subsection{Main Results on Exemplar CoT variant}
While this work does not investigates effectiveness of exemplar Chain of Thought as presented in RQ1, we do however test the main results of the work with Exemplar CoT to identify whether our findings hold true there as well. That is, we test \texttt{RQ3-Both, RQ3-One}. For \texttt{GPT-3.5-Turbo} we find that the average performance drops from 46.6\% (RQ3-Exemplar CoT) as in Table \ref{table:h1-h2} to 28.3\% in \texttt{One} and 10.4\% in \texttt{Both} variation cases, and remains at 40.3\% for \texttt{All} variation.

\section{Experiment Design}

Each of the variations proposed along RQ1, RQ2 and RQ3 modifies the few-shot examples only. Remaining aspects such as the query problem or the interaction with the simulator is directly inherited from the ReAct code-base \cite{yao2022react} at publicly available at \url{https://github.com/ysymyth/ReAct}. Our code can be found in the attached supplementary material.

Except \texttt{All} RQ3 variation, all other settings use the standard two examplars for prompting the LLM. Depending on the variation we change the content of the exemplar. Full prompts can be found in the attached supplementary code.

\subsection{Running the experiments}

In our experiments, according to the variation style we take the exemplar prompts and use the same exemplar prompts across the instances of the query task. Other than \texttt{RQ3-Both, One} we use the exemplar of the same task as the query as done in ReAct (and still find brittleness of ReAct). For \texttt{RQ3 - Both, One} we use exactly two exemplars but of a different task than query. Finally, \texttt{RQ3-All} is the only variation that provides six exemplars (instead of two) and we force the exemplar of the query-task to be appended at the end in the prompt. This was the best performing prompting strategy (on GPT-3.5-Turbo) amongst when the query-task exemplar was placed at the beginning, at position 4 (middle) and at the end. 

\subsection{Hyperparameters}
We use $temperature = \tau = 0$ for all of the GPT and Claude models and set \texttt{max-tokens = 100} which is borrowed from ReAct's hyperparameters. Rest of the parameters are kept to be default as specified in the respective model's API documentation.

\section{Example Prompts}
The full list of curated variations can be found in the supplementary materials. However, for completeness we reference the prompt used for base ReAct (as in \citep{yao2022react}) and our variations along RQ1, RQ2 and RQ3 for the \texttt{Put} task. 

\makeatletter
\def\tcb@proc@counter@auto#1{%
  \newcounter{tcb@cnt@#1}%
  \csxdef{tcb@cnt@#1}{tcb@cnt@#1}%
  \tcb@proc@counter@autoanduse{#1}%
  \ifcsname resetcounteronoverlays\endcsname
  \resetcounteronoverlays{tcb@cnt@#1}
  \fi
}
\makeatother
\newtcolorbox[auto counter]{numberedbox}[2][]{%
colback=lightgray!5,colframe=gray!40!black,center,title=Prompt ~\thetcbcounter: #2,#1, left skip=0pt, right skip=0pt, width=\linewidth}

\subsection{Synonym Substitution mapping for \texttt{Domain}}
We make the following substitutions to the object names  / locations in the exemplar prompt in the \texttt{Domain} variation. Note that these substitutions are done only to the exemplar, and the query problem and subsequent interaction with the AlfWorld simulator uses the original vocabulary mapping. 

\begin{verbatim}
spraybottle -> aerosolbottle
cabinet -> cupboard
countertop -> worktop
sinkbasin -> sinkbowl
toilet -> lavatory
toiletpaperhanger -> toiletpaperholder
towelholder -> towelrack

microwave -> oven 
shelf -> rack
drawer -> deskdrawer 
stoveburner -> hob
diningtable -> table
garbagecan -> trashbin

fridge -> refrigerator
peppershaker -> pepperpot
room -> livingroom
bread -> breadloaf 
pan -> fryingpan
pot -> saucepan
book -> notebook
 
creditcard -> amexcard 
mirror -> lookingglass
dresser -> chestofdrawers
sofa -> couch
cellphone -> mobilephone
coffeemachine -> coffeemaker
knife -> kitchenknife
spatula -> turner
soapbottle -> liquidsoapdispenser
saltshaker -> saltpot
statue -> sculpture
vase -> flowerpot
dishsponge -> spongewipe
desklamp -> tablelamp
sidetable -> nightstand
\end{verbatim}

\subsection{For \texttt{All, Both, One}}
\textbf{All}: We take the exemplar prompt for each task and concatenate it together.

\textbf{Both} : We use the following mapping generated arbitrarily, to replace the exemplar prompt for the query task. 
\begin{verbatim}
VARIATION_MAPPING = {
    'put':'examine',
    'clean':'cool',
    'heat':'put',
    'cool':'puttwo',
    'examine':'clean',
    'puttwo':'heat'
}
\end{verbatim}

\textbf{One} : We use the first example from the above mapping, and the second example from the same task as the query. 

\subsection{Building the input prompt}
\label{app:building-prompt}
We inherit ReAct's way of stiching together the exemplars and the query using the following prompt template as : 

\texttt{Interact with a household to solve a task. Here are two examples.' + eg1 + eg2 + INSTRUCTION + Here is the task. (TASK)\'}

In the case of GPT models \texttt{INSTRUCTION} is an empty string. Claude Models gave 0\% success rate (regardless of the model size, i.e. Haiku, Sonnet and Opus) with the original ReAct prompt design. We identified that Claude required an explicit instruction to generate only the actions for the simulator. We added a task independent INSTRUCTION prompt as : 

\texttt{INSTRUCTION = " Only provide the one action at a time. Be concise, and do not provide any extra information. Always start with the action. For example, 'pick up apple' is correct, but 'I want to pick up the apple' is not."}

Interestingly, GPT-3-Turbo performance dropped to 1\% (averaged across 6 independent runs over 134 examples) with the inclusion of this INSTRUCTION tag. This further highlights how brittle prompt engineering is, and cautions us against misplacing our trust with ReAct in enhancing reasoning abilities of LLMs.

\subsection{RQ1}
    \begin{numberedbox}[label={  prompt:rq1-anon-exemplar  }]{ rq1-anon-exemplar }
    {\scriptsize
    You are in the middle of a room. Looking quickly around you, you see a cabinet 4, a cabinet 3, a cabinet 2, a cabinet 1, a coffeemachine 1, a countertop 1, a diningtable 3, a diningtable 2, a diningtable 1, a drawer 1, a fridge 1, a garbagecan 1, a microwave 1, a sidetable 1, a sinkbasin 1, a stoveburner 4, a stoveburner 3, a stoveburner 2, a stoveburner 1, and a toaster 1.

Your task is to: find some apple and put it in sidetable.

> think: To solve the task, I need to find and take the object, then put it on the desired location. First I need to find the object. The object is more likely to appear in one of the different locations. I can check one by one, starting with the first location. Once I find the object, next I need to take it. Once I take the object, next I need to put it in/on the desired location.

OK.

> go to fridge 1

The fridge 1 is closed.

> open fridge 1

You open the fridge 1. The fridge 1 is open. In it, you see a lettuce 2, a mug 2, a potato 2, and a tomato 1.

> go to diningtable 1

On the diningtable 1, you see nothing.

> go to diningtable 2

On the diningtable 2, you see a pineapple 1.

> go to diningtable 3

On the diningtable 3, you see a bowl 2, a bowl 1, a bread 2, and a bread 1.

> go to sidetable 1

On the sidetable 1, you see a cup 1, a lettuce 1, a peppershaker 3, a potato 1, and a saltshaker 1.

> go to countertop 1

On the countertop 1, you see a butterknife 1, a dishsponge 1, and a soapbottle 1.

> go to sinkbasin 1

On the sinkbasin 1, you see a cup 2, a egg 1, a mug 1, and a potato 3.

> go to garbagecan 1

On the garbagecan 1, you see a apple 3, and a egg 3.

> take apple 3 from garbagecan 1

You pick up the apple 3 from the garbagecan 1.

> go to sidetable 1

On the sidetable 1, you see a cup 1, a lettuce 1, a peppershaker 3, a potato 1, and a saltshaker 1.

> put apple 3 in/on sidetable 1

You put the apple 3 in/on the sidetable 1.

    }
    \end{numberedbox}

    \begin{numberedbox}[label={  prompt:rq1-exemplar  }]{ rq1-exemplar }
    {\scriptsize
    You are in the middle of a room. Looking quickly around you, you see a cabinet 4, a cabinet 3, a cabinet 2, a cabinet 1, a coffeemachine 1, a countertop 1, a diningtable 3, a diningtable 2, a diningtable 1, a drawer 1, a fridge 1, a garbagecan 1, a microwave 1, a sidetable 1, a sinkbasin 1, a stoveburner 4, a stoveburner 3, a stoveburner 2, a stoveburner 1, and a toaster 1.

Your task is to: find some apple and put it in sidetable.

> think: To solve the task, I need to find and take an apple, then put it in sidetable. First I need to find an apple. An apple is more likely to appear in fridges (1), diningtables (1-3), sidetables (1), countertops (1), sinkbasins (1), garbagecan (1). I can check one by one, starting with fridge 1. Once I find an apple, next I need to take it. Once I find an apple, next I need to put it in/on sidetable.

OK.

> go to fridge 1

The fridge 1 is closed.

> open fridge 1

You open the fridge 1. The fridge 1 is open. In it, you see a lettuce 2, a mug 2, a potato 2, and a tomato 1.

> go to diningtable 1

On the diningtable 1, you see nothing.

> go to diningtable 2

On the diningtable 2, you see a pineapple 1.

> go to diningtable 3

On the diningtable 3, you see a bowl 2, a bowl 1, a bread 2, and a bread 1.

> go to sidetable 1

On the sidetable 1, you see a cup 1, a lettuce 1, a peppershaker 3, a potato 1, and a saltshaker 1.

> go to countertop 1

On the countertop 1, you see a butterknife 1, a dishsponge 1, and a soapbottle 1.

> go to sinkbasin 1

On the sinkbasin 1, you see a cup 2, a egg 1, a mug 1, and a potato 3.

> go to garbagecan 1

On the garbagecan 1, you see a apple 3, and a egg 3.

> take apple 3 from garbagecan 1

You pick up the apple 3 from the garbagecan 1.

> go to sidetable 1

On the sidetable 1, you see a cup 1, a lettuce 1, a peppershaker 3, a potato 1, and a saltshaker 1.

> put apple 3 in/on sidetable 1

You put the apple 3 in/on the sidetable 1.

    }
    \end{numberedbox}

    \begin{numberedbox}[label={  prompt:base-react  }]{ base-react }
    {\scriptsize
    You are in the middle of a room. Looking quickly around you, you see a cabinet 4, a cabinet 3, a cabinet 2, a cabinet 1, a coffeemachine 1, a countertop 1, a diningtable 3, a diningtable 2, a diningtable 1, a drawer 1, a fridge 1, a garbagecan 1, a microwave 1, a sidetable 1, a sinkbasin 1, a stoveburner 4, a stoveburner 3, a stoveburner 2, a stoveburner 1, and a toaster 1.

Your task is to: find some apple and put it in sidetable.

> think: First I need to find an apple. An apple is more likely to appear in fridges (1), diningtables (1-3), sidetables (1), countertops (1), sinkbasins (1), garbagecan (1). I can check one by one, starting with fridge 1.

OK.

> go to fridge 1

The fridge 1 is closed.

> open fridge 1

You open the fridge 1. The fridge 1 is open. In it, you see a lettuce 2, a mug 2, a potato 2, and a tomato 1.

> go to diningtable 1

On the diningtable 1, you see nothing.

> go to diningtable 2

On the diningtable 2, you see a pineapple 1.

> go to diningtable 3

On the diningtable 3, you see a bowl 2, a bowl 1, a bread 2, and a bread 1.

> go to sidetable 1

On the sidetable 1, you see a cup 1, a lettuce 1, a peppershaker 3, a potato 1, and a saltshaker 1.

> go to countertop 1

On the countertop 1, you see a butterknife 1, a dishsponge 1, and a soapbottle 1.

> go to sinkbasin 1

On the sinkbasin 1, you see a cup 2, a egg 1, a mug 1, and a potato 3.

> go to garbagecan 1

On the garbagecan 1, you see a apple 3, and a egg 3.

> take apple 3 from garbagecan 1

You pick up the apple 3 from the garbagecan 1.

> think: Now I find a apple (3). Next, I need to put it in/on sidetable 1.

OK.

> go to sidetable 1

On the sidetable 1, you see a cup 1, a lettuce 1, a peppershaker 3, a potato 1, and a saltshaker 1.

> put apple 3 in/on sidetable 1

You put the apple 3 in/on the sidetable 1.

    }
    \end{numberedbox}

\newpage

\subsection{RQ2}
    \begin{numberedbox}[label={  prompt:rq2-Explanation  }]{ rq2-Explanation }
    {\scriptsize
    You are in the middle of a room. Looking quickly around you, you see a cabinet 4, a cabinet 3, a cabinet 2, a cabinet 1, a coffeemachine 1, a countertop 1, a diningtable 3, a diningtable 2, a diningtable 1, a drawer 1, a fridge 1, a garbagecan 1, a microwave 1, a sidetable 1, a sinkbasin 1, a stoveburner 4, a stoveburner 3, a stoveburner 2, a stoveburner 1, and a toaster 1.

Your task is to: find some apple and put it in sidetable.

> go to fridge 1

The fridge 1 is closed.

> close cabinet 2.

Nothing Happens.

> think: Nothing happens because you are not at cabinet 2.

OK.

> open fridge 1

You open the fridge 1. The fridge 1 is open. In it, you see a lettuce 2, a mug 2, a potato 2, and a tomato 1.

> go to diningtable 1

On the diningtable 1, you see nothing.

> go to diningtable 2

On the diningtable 2, you see a pineapple 1.

> go to diningtable 3

On the diningtable 3, you see a bowl 2, a bowl 1, a bread 2, and a bread 1.

> go to sidetable 1

On the sidetable 1, you see a cup 1, a lettuce 1, a peppershaker 3, a potato 1, and a saltshaker 1.

> go to countertop 1

On the countertop 1, you see a butterknife 1, a dishsponge 1, and a soapbottle 1.

> go to sinkbasin 1

On the sinkbasin 1, you see a cup 2, a egg 1, a mug 1, and a potato 3.

> go to garbagecan 1

On the garbagecan 1, you see a apple 3, and a egg 3.

> put apple 3 in/on sidetable.

Nothing Happens.

> think: Nothing happens because I do not have apple 3.

OK.

> take apple 3 from garbagecan 1

You pick up the apple 3 from the garbagecan 1.

> go to sidetable 1

On the sidetable 1, you see a cup 1, a lettuce 1, a peppershaker 3, a potato 1, and a saltshaker 1.

> put apple 3 in/on sidetable 1

You put the apple 3 in/on the sidetable 1.

    }
    \end{numberedbox}

    \begin{numberedbox}[label={  prompt:rq2-ordering  }]{ rq2-ordering }
    {\scriptsize
    You are in the middle of a room. Looking quickly around you, you see a cabinet 4, a cabinet 3, a cabinet 2, a cabinet 1, a coffeemachine 1, a countertop 1, a diningtable 3, a diningtable 2, a diningtable 1, a drawer 1, a fridge 1, a garbagecan 1, a microwave 1, a sidetable 1, a sinkbasin 1, a stoveburner 4, a stoveburner 3, a stoveburner 2, a stoveburner 1, and a toaster 1.

Your task is to: find some apple and put it in sidetable.

> think: To solve the task, I need to put an apple in sidetable, for which I need to find and take an apple.

OK.

> think: An apple is more likely to appear in sidetables (1), fridges (1), countertops (1), diningtables (1-3), garbagecan (1), sinkbasins (1). First I need to find an apple. I can check one by one, starting with fridge 1.

OK.

> go to fridge 1

The fridge 1 is closed.

> open fridge 1

You open the fridge 1. The fridge 1 is open. In it, you see a lettuce 2, a mug 2, a potato 2, and a tomato 1.

> go to diningtable 1

On the diningtable 1, you see nothing.

> go to diningtable 2

On the diningtable 2, you see a pineapple 1.

> go to diningtable 3

On the diningtable 3, you see a bowl 2, a bowl 1, a bread 2, and a bread 1.

> go to sidetable 1

On the sidetable 1, you see a cup 1, a lettuce 1, a peppershaker 3, a potato 1, and a saltshaker 1.

> go to countertop 1

On the countertop 1, you see a butterknife 1, a dishsponge 1, and a soapbottle 1.

> go to sinkbasin 1

On the sinkbasin 1, you see a cup 2, a egg 1, a mug 1, and a potato 3.

> go to garbagecan 1

On the garbagecan 1, you see a apple 3, and a egg 3.

> think: Next, I need to take the apple.Now I find an apple (3).

OK.

> take apple 3 from garbagecan 1

You pick up the apple 3 from the garbagecan 1.

> think: Next, I need to put the apple in/on sidetable 1. Now I find an apple (3). 

OK.

> go to sidetable 1

On the sidetable 1, you see a cup 1, a lettuce 1, a peppershaker 3, a potato 1, and a saltshaker 1.

> put apple 3 in/on sidetable 1

You put the apple 3 in/on the sidetable 1.

    }
    \end{numberedbox}

    \begin{numberedbox}[label={  prompt:rq2-Placebo  }]{ rq2-Placebo }
    {\scriptsize
    You are in the middle of a room. Looking quickly around you, you see a cabinet 4, a cabinet 3, a cabinet 2, a cabinet 1, a coffeemachine 1, a countertop 1, a diningtable 3, a diningtable 2, a diningtable 1, a drawer 1, a fridge 1, a garbagecan 1, a microwave 1, a sidetable 1, a sinkbasin 1, a stoveburner 4, a stoveburner 3, a stoveburner 2, a stoveburner 1, and a toaster 1.

Your task is to: find some apple and put it in sidetable.

> think: Take a deep breath and work on this problem step-by-step.

OK.

> go to fridge 1

The fridge 1 is closed.

> open fridge 1

You open the fridge 1. The fridge 1 is open. In it, you see a lettuce 2, a mug 2, a potato 2, and a tomato 1.

> go to diningtable 1

On the diningtable 1, you see nothing.

> go to diningtable 2

On the diningtable 2, you see a pineapple 1.

> go to diningtable 3

On the diningtable 3, you see a bowl 2, a bowl 1, a bread 2, and a bread 1.

> go to sidetable 1

On the sidetable 1, you see a cup 1, a lettuce 1, a peppershaker 3, a potato 1, and a saltshaker 1.

> go to countertop 1

On the countertop 1, you see a butterknife 1, a dishsponge 1, and a soapbottle 1.

> go to sinkbasin 1

On the sinkbasin 1, you see a cup 2, a egg 1, a mug 1, and a potato 3.

> go to garbagecan 1

On the garbagecan 1, you see a apple 3, and a egg 3.

> think: Take a deep breath and work on this problem step-by-step.

OK.

> take apple 3 from garbagecan 1

You pick up the apple 3 from the garbagecan 1.

> think: Take a deep breath and work on this problem step-by-step.

OK.

> go to sidetable 1

On the sidetable 1, you see a cup 1, a lettuce 1, a peppershaker 3, a potato 1, and a saltshaker 1.

> put apple 3 in/on sidetable 1

You put the apple 3 in/on the sidetable 1.

    }
    \end{numberedbox}

    \begin{numberedbox}[label={  prompt:rq2-Failure  }]{ rq2-Failure }
    {\scriptsize
    You are in the middle of a room. Looking quickly around you, you see a cabinet 4, a cabinet 3, a cabinet 2, a cabinet 1, a coffeemachine 1, a countertop 1, a diningtable 3, a diningtable 2, a diningtable 1, a drawer 1, a fridge 1, a garbagecan 1, a microwave 1, a sidetable 1, a sinkbasin 1, a stoveburner 4, a stoveburner 3, a stoveburner 2, a stoveburner 1, and a toaster 1.

Your task is to: find some apple and put it in sidetable.

> go to fridge 1

The fridge 1 is closed.

> close cabinet 2.

Nothing Happens.

> open fridge 1

You open the fridge 1. The fridge 1 is open. In it, you see a lettuce 2, a mug 2, a potato 2, and a tomato 1.

> go to diningtable 1

On the diningtable 1, you see nothing.

> go to diningtable 2

On the diningtable 2, you see a pineapple 1.

> go to diningtable 3

On the diningtable 3, you see a bowl 2, a bowl 1, a bread 2, and a bread 1.

> go to sidetable 1

On the sidetable 1, you see a cup 1, a lettuce 1, a peppershaker 3, a potato 1, and a saltshaker 1.

> go to countertop 1

On the countertop 1, you see a butterknife 1, a dishsponge 1, and a soapbottle 1.

> go to sinkbasin 1

On the sinkbasin 1, you see a cup 2, a egg 1, a mug 1, and a potato 3.

> go to garbagecan 1

On the garbagecan 1, you see a apple 3, and a egg 3.

> put apple 3 in/on sidetable.

Nothing Happens.

> take apple 3 from garbagecan 1

You pick up the apple 3 from the garbagecan 1.

> go to sidetable 1

On the sidetable 1, you see a cup 1, a lettuce 1, a peppershaker 3, a potato 1, and a saltshaker 1.

> put apple 3 in/on sidetable 1

You put the apple 3 in/on the sidetable 1.

    }
    \end{numberedbox}

\newpage 

\subsection{RQ3}
    \begin{numberedbox}[label={  prompt:rq3-Domain  }]{ rq3-Domain }
    {\scriptsize
    You are in the middle of a livingroom. Looking quickly around you, you see a cupboard 4, a cupboard 3, a cupboard 2, a cupboard 1, a coffeemaker 1, a worktop 1, a table 3, a table 2, a table 1, a deskdrawer 1, a refrigerator 1, a trashbin 1, a oven 1, a nightstand 1, a sinkbowl 1, a hob 4, a hob 3, a hob 2, a hob 1, and a toaster 1.

Your task is to: find some apple and put it in nightstand.

> think: To solve the task, I need to find and take an apple, then put it in nightstand.

OK.

> think: First I need to find an apple. An apple is more likely to appear in refrigerators (1), tables (1-3), nightstands (1), worktops (1), sinkbowls (1), trashbin (1). I can check one by one, starting with refrigerator 1.

OK.

> go to refrigerator 1

The refrigerator 1 is closed.

> oballpointpen refrigerator 1

You oballpointpen the refrigerator 1. The refrigerator 1 is oballpointpen. In it, you see a lettuce 2, a mug 2, a saucepanato 2, and a tomato 1.

> go to table 1

On the table 1, you see nothing.

> go to table 2

On the table 2, you see a pineapple 1.

> go to table 3

On the table 3, you see a bowl 2, a bowl 1, a breadloaf 2, and a breadloaf 1.

> go to nightstand 1

On the nightstand 1, you see a cup 1, a lettuce 1, a peppersaucepan 3, a saucepanato 1, and a saltpot 1.

> go to worktop 1

On the worktop 1, you see a butterknife 1, a spongewipe 1, and a liquidsoapdispenser 1.

> go to sinkbowl 1

On the sinkbowl 1, you see a cup 2, a egg 1, a mug 1, and a saucepanato 3.

> go to trashbin 1

On the trashbin 1, you see a apple 3, and a egg 3.

> think: Now I find an apple (3). Next, I need to take it.

> take apple 3 from trashbin 1

You pick up the apple 3 from the trashbin 1.

> think: Now I find an apple (3). Next, I need to put it in/on nightstand 1.

OK.

> go to nightstand 1

On the nightstand 1, you see a cup 1, a lettuce 1, a peppersaucepan 3, a saucepanato 1, and a saltpot 1.

> put apple 3 in/on nightstand 1

You put the apple 3 in/on the nightstand 1.

    }
    \end{numberedbox}

    \begin{numberedbox}[label={  prompt:rq3-Instance  }]{ rq3-Instance }
    {\scriptsize
    You are in the middle of a room. Looking quickly around you, you see a cabinet 4, a cabinet 3, a cabinet 2, a cabinet 1, a coffeemachine 1, a countertop 1, a diningtable 3, a diningtable 2, a diningtable 1, a drawer 1, a fridge 1, a garbagecan 1, a microwave 1, a sidetable 1, a sinkbasin 1, a stoveburner 4, a stoveburner 3, a stoveburner 2, a stoveburner 1, and a toaster 1.

Your task is to: find some apple and put it in sidetable.

> think: To solve the task, I need to find and take an apple, then put it in sidetable.

OK.

> think: First I need to find an apple. An apple is more likely to appear in fridges (1), diningtables (1-3), sidetables (1), countertops (1), sinkbasins (1), garbagecan (1). I can check one by one, starting with fridge 1.

OK.

> go to fridge 1

The fridge 1 is closed.

> open fridge 1

You open the fridge 1. The fridge 1 is open. In it, you see a lettuce 2, a mug 2, a potato 2, and a tomato 1.

> go to diningtable 1

On the diningtable 1, you see nothing.

> go to diningtable 2

On the diningtable 2, you see a pineapple 1.

> go to diningtable 3

On the diningtable 3, you see a bowl 2, a bowl 1, a bread 2, and a bread 1.

> go to sidetable 1

On the sidetable 1, you see a cup 1, a lettuce 1, a peppershaker 3, a potato 1, and a saltshaker 1.

> go to countertop 1

On the countertop 1, you see a butterknife 1, a dishsponge 1, and a soapbottle 1.

> go to sinkbasin 1

On the sinkbasin 1, you see a cup 2, a egg 1, a mug 1, and a potato 3.

> go to diningtable 2

On the diningtable 2, you see a pineapple 1.

> go to diningtable 3

On the diningtable 3, you see a bowl 2, a bowl 1, a bread 2, and a bread 1.

> go to sidetable 1

On the sidetable 1, you see a cup 1, a lettuce 1, a peppershaker 3, a potato 1, and a saltshaker 1.

> go to countertop 1

On the countertop 1, you see a butterknife 1, a dishsponge 1, and a soapbottle 1.

> go to sinkbasin 1

On the sinkbasin 1, you see a cup 2, a egg 1, a mug 1, and a potato 3.

> go to garbagecan 1

On the garbagecan 1, you see a apple 3, and a egg 3.

> think: Now I find an apple (3). Next, I need to take it.

> take apple 3 from garbagecan 1

You pick up the apple 3 from the garbagecan 1.

> think: Now I find an apple (3). Next, I need to put it in/on sidetable 1.

OK.

> go to sidetable 1

On the sidetable 1, you see a cup 1, a lettuce 1, a peppershaker 3, a potato 1, and a saltshaker 1.

> put apple 3 in/on sidetable 1

You put the apple 3 in/on the sidetable 1.

    }
    \end{numberedbox}

    \begin{numberedbox}[label={  prompt:rq3-Optimal  }]{ rq3-Optimal }
    {\scriptsize
    You are in the middle of a room. Looking quickly around you, you see a cabinet 4, a cabinet 3, a cabinet 2, a cabinet 1, a coffeemachine 1, a countertop 1, a diningtable 3, a diningtable 2, a diningtable 1, a drawer 1, a fridge 1, a garbagecan 1, a microwave 1, a sidetable 1, a sinkbasin 1, a stoveburner 4, a stoveburner 3, a stoveburner 2, a stoveburner 1, and a toaster 1.

Your task is to: find some apple and put it in sidetable.

> think: To solve the task, I need to find and take an apple, then put it in sidetable.

OK.

> think: First I need to find an apple. An apple is more likely to appear in fridges (1), diningtables (1-3), sidetables (1), countertops (1), sinkbasins (1), garbagecan (1). I can check one by one, starting with garbagecan 1.

OK.

> go to garbagecan 1

On the garbagecan 1, you see a apple 3, and a egg 3.

> think: Now I find an apple (3). Next, I need to take it.

> take apple 3 from garbagecan 1

You pick up the apple 3 from the garbagecan 1.

> think: Now I find an apple (3). Next, I need to put it in/on sidetable 1.

OK.

> go to sidetable 1

On the sidetable 1, you see a cup 1, a lettuce 1, a peppershaker 3, a potato 1, and a saltshaker 1.

> put apple 3 in/on sidetable 1

You put the apple 3 in/on the sidetable 1.

    }
    \end{numberedbox}

\newpage

\end{document}